\documentclass{article}

\usepackage{microtype}
\usepackage{graphicx}
\usepackage{subfigure}
\usepackage{booktabs} 

\usepackage[hidelinks]{hyperref}

\usepackage{icml2025}

\usepackage{amsmath}
\usepackage{amssymb}
\usepackage{mathtools}
\usepackage{amsthm}
\usepackage[T1]{fontenc}
\usepackage[utf8]{inputenc}
\usepackage{listings}

\usepackage{color}

\usepackage[capitalize,noabbrev]{cleveref}

\theoremstyle{plain}

\theoremstyle{definition}

\theoremstyle{remark}

\usepackage[textsize=tiny]{todonotes}

\allowdisplaybreaks

\definecolor{keywordcolor}{rgb}{0.7, 0.1, 0.1}   
\definecolor{tacticcolor}{rgb}{0.0, 0.1, 0.6}    
\definecolor{commentcolor}{rgb}{0.4, 0.4, 0.4}   
\definecolor{symbolcolor}{rgb}{0.0, 0.1, 0.6}    
\definecolor{sortcolor}{rgb}{0.1, 0.5, 0.1}      
\definecolor{attributecolor}{rgb}{0.7, 0.1, 0.1} 

\icmltitlerunning{LeanConjecturer: Automatic Generation of Mathematical Conjectures for Theorem Proving}

\begin{document}

\twocolumn[
\icmltitle{LeanConjecturer: Automatic Generation of Mathematical Conjectures for Theorem Proving}



\icmlsetsymbol{equal}{*}


\def\isaccepted{}
\begin{icmlauthorlist}
\icmlauthor{Naoto Onda}{osx,nexa,autores}
\icmlauthor{Kazumi Kasaura}{osx,autores}
\icmlauthor{Yuta Oriike}{ca,autores}
\icmlauthor{Masaya Taniguchi}{aip,autores}
\icmlauthor{Akiyoshi Sannai}{kyoto,trip,shiga,nii,nistep,autores}
\icmlauthor{Sho Sonoda}{aip,ca,autores}
\end{icmlauthorlist}

\icmlaffiliation{osx}{OMRON SINIC X Corporation}
\icmlaffiliation{nexa}{NexaScience}
\icmlaffiliation{aip}{RIKEN AIP}
\icmlaffiliation{trip}{Advanced General Intelligence for Science Program, Kobe, Japan}
\icmlaffiliation{kyoto}{Department of Physics, Kyoto University}
\icmlaffiliation{shiga}{Data Science and AI Innovation Research Promotion Center, Shiga University}
\icmlaffiliation{nii}{Research and Development Center for Large Language Models, National Institute of Informatics}
\icmlaffiliation{nistep}{National Institute of Science and Technology Policy (NISTEP)}
\icmlaffiliation{ca}{CyberAgent}
\icmlaffiliation{autores}{AutoRes}

\icmlcorrespondingauthor{Naoto Onda}{naoto.onda@sinicx.com}

\icmlkeywords{theorem generating, hypothesis generation, theorem proving, Lean, LLMs}

\vskip 0.3in
]



\printAffiliationsAndNotice{}  

\begin{abstract}
We introduce LeanConjecturer, a pipeline for automatically %
generating university-level mathematical conjectures in Lean 4 %
using Large Language Models (LLMs). %
Our hybrid approach combines rule-based context extraction with LLM-based theorem statement generation, %
addressing the data scarcity challenge in formal theorem proving. %
Through iterative generation and evaluation, LeanConjecturer produced 12,289 conjectures from 40 Mathlib seed files, %
with 3,776 identified as syntactically valid and non-trivial, that is, cannot be proven by \texttt{aesop} tactic. %
We demonstrate the utility of these generated conjectures for reinforcement learning %
through Group Relative Policy Optimization (GRPO), %
showing that targeted training on domain-specific conjectures can enhance theorem proving capabilities. %
Our approach generates 103.25 novel conjectures per seed file on average, %
providing a scalable solution for creating training data for theorem proving systems. %
Our system successfully verified several non-trivial theorems in topology, %
including properties of semi-open, alpha-open, and pre-open sets, %
demonstrating its potential for mathematical discovery beyond simple variations of existing results.
\end{abstract}

\section{Introduction}
Large Language Models (LLMs) have demonstrated significant promise in the field of formal theorem proving, %
a critical area for ensuring the reliability and reasoning capabilities of AI systems. %
This burgeoning field aims to enable machines to construct and verify mathematical proofs %
using interactive theorem provers (ITPs) such as Lean\footnote{\url{https://leanprover.github.io/}}, %
thereby enhancing the trustworthiness of LLM outputs in complex logical tasks \cite{liu2025safeenhancingmathematicalreasoning}. %
Recent advancements underscore this potential, with projects such as DeepSeek-Prover-V2 \cite{ren2025deepseekproverv2advancingformalmathematical} %
achieving state-of-the-art performance on rigorous formal mathematics benchmarks. %
For instance, DeepSeek-Prover-V2 has attained an impressive 88.9\% pass ratio on the MiniF2F-test benchmark \cite{conf/iclr/ZhengHP22} %
and successfully solved 49 out of 658 problems from PutnamBench \cite{tsoukalas2024putnambenchevaluatingneuraltheoremprovers}, %
further demonstrating its sophisticated mathematical reasoning by tackling challenging AIME problems. %

Despite these advancements, a key bottleneck in applying LLMs to Lean is the inherent scarcity of high-quality training data in formal formats. %
Unlike the vast and readily available corpora of informal mathematics, which span terabytes of data typically used for general LLM pre-training, %
Lean's mathematical library (Mathlib\footnote{\url{https://github.com/leanprover-community/mathlib}}) and existing benchmarks %
such as MiniF2F provide a relatively limited number of example theorems and proofs for supervised training. %
Quantitatively, standard Lean 4 libraries, including Mathlib, collectively less than 1GB of data, %
which is insufficient for effectively fine-tuning LLMs that thrive on extensive data exposure. %
The process of formalizing informal mathematical reasoning into machine-verifiable proofs is also labor-intensive and time-consuming, %
demanding extreme precision at every logical step, even for experienced mathematicians. %

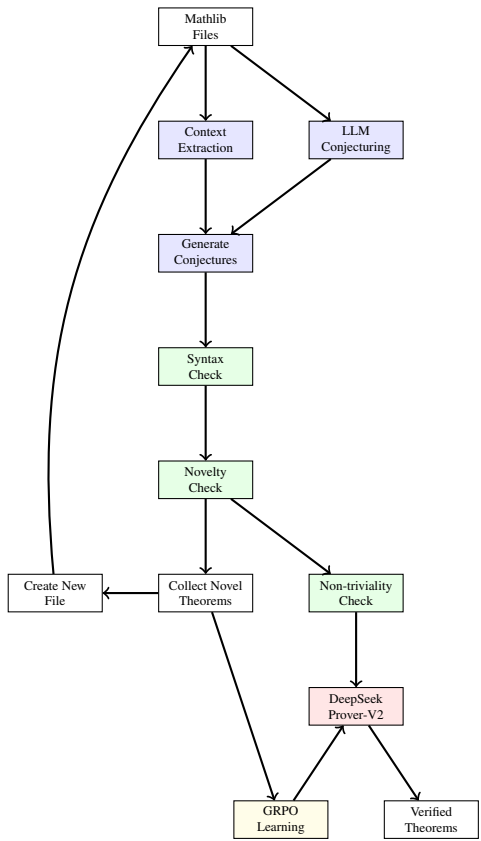
\begin{figure}[h]
    \centering
    \begin{tikzpicture}[
        node distance=2cm,
        box/.style={draw, rectangle, minimum width=2.5cm, minimum height=1cm, align=center},
        arrow/.style={->, thick},
        process/.style={box, fill=blue!10},
        eval/.style={box, fill=green!10},
        verify/.style={box, fill=red!10},
        learn/.style={box, fill=yellow!10},
        scale=0.5, transform shape
    ]
        \node[box] (input) {Mathlib\\Files};

        \node[process] (context) [below=of input] {Context\\Extraction};
        \node[process] (llm) [below=of input, xshift=4cm] {LLM\\Conjecturing};
        \node[process] (conjecture) [below=of context] {Generate\\Conjectures};

        \node[eval] (syntax) [below=of conjecture] {Syntax\\Check};
        \node[eval] (novelty) [below=of syntax] {Novelty\\Check};
        \node[eval] (nontriv) [below=of novelty, xshift=4cm] {Non-triviality\\Check};

        \node[box] (collect) [below=of novelty] {Collect Novel\\Theorems};
        \node[box] (newfile) [below=of novelty, xshift=-4cm] {Create New\\File};

        \node[verify] (prover) [below=of nontriv] {DeepSeek\\Prover-V2};
        \node[learn] (grpo) [below=of prover, xshift=-2cm] {GRPO\\Learning};
        \node[box] (verified) [below=of prover, xshift=2cm] {Verified\\Theorems};

        \draw[arrow] (input) -- (llm);
        \draw[arrow] (input) -- (context);
        \draw[arrow] (context) -- (conjecture);
        \draw[arrow] (llm) -- (conjecture);
        \draw[arrow] (conjecture) -- (syntax);
        \draw[arrow] (syntax) -- (novelty);
        \draw[arrow] (novelty) -- (nontriv);
        \draw[arrow] (novelty) -- (collect);
        \draw[arrow] (collect) -- (newfile);
        \draw[arrow] (newfile) to[bend left=20] (input);
        \draw[arrow] (nontriv) -- (prover);
        \draw[arrow] (collect) -- (grpo);
        \draw[arrow] (grpo) -- (prover);
        \draw[arrow] (prover) -- (verified);
    \end{tikzpicture}
    \caption{Overview of our theorem generation and verification pipeline. The process starts with Mathlib files as input, goes through generation (blue), evaluation (green), and iteration phases, and culminates in verification (red) and reinforcement learning (yellow). The iterative nature of the pipeline allows for continuous generation of novel conjectures while maintaining mathematical validity.}
    \label{fig:pipeline_overview}
\end{figure}

To address this challenge and overcome the ``triviality trap'' of generating syntactically correct %
but mathematically uninteresting propositions, we propose a novel pipeline for automatically generating mathematical conjectures %
in Lean 4 format (see Figure~\ref{fig:pipeline_overview}). %
We call this pipeline LeanConjecturer. %
LeanConjecturer strategically leverages existing Mathlib files as inspiration. %
It employs a hybrid strategy that combines rule-based context extraction with LLM-based theorem statement generation, %
enabling more flexible and cost-effective generation of mathematical statements without requiring existing proofs. %

We demonstrate that our generated conjectures can be utilized for reinforcement learning %
through Group Relative Policy Optimization (GRPO), improving theorem proving capabilities %
particularly in domains such as topology and set theory. %
This integrated approach pushes the boundaries of AI in mathematics, moving towards true mathematical discovery.

Our implementation, including the conjecture generation pipeline, evaluation scripts, and reinforcement learning setup, %
is publicly available at \url{https://github.com/auto-res/LeanConjecturer.git} and \url{https://github.com/auto-res/open-r1.git} for reproducibility and further research.

\section{Related Work}

\subsection{Formal Theorem Proving with LLMs}
Recent years have witnessed remarkable progress in applying LLMs to formal theorem proving. %
This remarkable progress is attributed to innovative strategies, including a theorem proving pipeline %
where DeepSeek-V3 \cite{deepseekai2025deepseekv3technicalreport} (671B model) performs subgoal decomposition %
and DeepSeek-Prover-V2 (7B model) handles the proof of subgoals. %
This pipeline facilitates the decomposition of complex problems into manageable subgoals, %
mirroring human problem-solving approaches and providing high-quality cold-start training data for reinforcement learning. %
Furthermore, tools known as ``hammers'' such as Lean-auto\footnote{\url{https://github.com/leanprover-community/lean-auto}}, %
have been developed to provide general-purpose proof automation in Lean 4, %
by translating between the ITP's logical system (dependent type theory in Lean 4's case) and Automated Theorem Provers (ATPs), %
ensuring soundness and completeness in practical applications.

\subsection{Data Scarcity and Autoformalization}
This data sparsity has led recent projects to cope by autoformalizing informal problems, %
particularly those from mathematics competitions such as the International Mathematical Olympiad (IMO) %
and platforms such as Art of Problem Solving (AOPS). %
Projects such as Lean Workbook \cite{ying2024internlmmathopenmathlarge} have demonstrated the effectiveness of autoformalizing %
mathematical problems from natural language, providing valuable training data for theorem proving systems. %
While this approach has yielded valuable training data and serves as a promising mechanism for grounding LLM reasoning in formal logic, %
it also presents limitations. %
Competition problems are typically short, self-contained statements that are easier for LLMs to parse and formalize, %
and they predominantly cover contest-level mathematics (e.g., olympiad geometry, number theory). %
This over-reliance on autoformalized contest data risks over-specializing current LLM theorem provers to a narrow style of problems, %
neglecting the broader spectrum of mathematics found in Lean's Mathlib, which includes more advanced and abstract theories. %

\subsection{Synthetic Data Generation}
Recent advances in synthetic data generation have demonstrated promising approaches for addressing data scarcity in formal mathematics. %
One notable technique is ``state graph exploration,'' exemplified by LeanNavigator \cite{yin2025generatingmillionsleantheorems}, %
which explores and extracts new provable states from Lean 4 proofs within Mathlib4. %
This approach constructs a state transition graph from existing proofs and systematically explores alternative proof paths, %
generating millions of theorems (4.7 million theorems, 1 billion tokens) by discovering new provable statements %
that can be derived through different proof strategies. %
LeanNavigator's state graph exploration methodology enables the discovery of alternative proofs %
and previously undiscovered mathematical relationships within the formal mathematical library, %
providing a valuable source of training data for theorem proving systems. %

\subsection{Self-Play of Conjecturers and Theorem Provers}

A recent breakthrough in addressing the data scarcity challenge is the introduction of self-play mechanisms in theorem proving. %
The Self-play Theorem Prover (STP) \cite{dong2025stpselfplayllmtheorem} represents a paradigm shift %
from static human-curated datasets to dynamic, self-generated curricula. %
Inspired by how human mathematicians develop new results, STP employs a dual-role approach %
where an LLM simultaneously acts as both a ``conjecturer'' and a ``prover,'' %
providing mutual training signals to each other. %
The conjecturer generates new conjectures that are ``barely provable'' for the current prover, %
creating an optimal challenge level that lies within the model's zone of proximal development. %
The prover then attempts to prove these generated conjectures using standard expert iteration, %
with the success or failure of proofs providing feedback to refine the conjecturer's ability to propose optimal learning tasks. %

This self-play mechanism addresses the fundamental limitation of many AI systems %
where performance plateaus after exhausting training data. %
By generating progressively challenging yet achievable learning tasks, %
STP enables continuous improvement without external data requirements. 

However, STP faces several challenges including mode collapse, %
where generated conjectures tend to focus on specific topics despite diverse seed statements, %
and training instability due to dynamic data distribution changes during self-play. %
The system also relies on heuristic ``elegance filters'' for conjecture selection, %
which may not fully capture mathematical elegance or relevance. %

This approach demonstrates the potential of self-generating curricula in theorem proving, %
offering a path toward open-ended learning and potentially AGI in complex reasoning domains. %
The success of self-play mechanisms suggests that AI systems can overcome data scarcity %
by creating their own challenging learning tasks, %
moving beyond the limitations of static training datasets. %

\section{Method}
\subsection{Generation}
To generate university-level mathematical conjectures using LLMs, %
we utilize files from Mathlib as inspiration, denoted by $X$. %
Rather than prompting an LLM directly to produce analogous conjectures and proofs, %
which often results in importing hallucinated libraries or grammatically incorrect theorems %
involving fictional lemmas, we adopt a structured generation process (see Figure~\ref{fig:pipeline_overview}). %
Specifically, we instruct the LLM to generate only the statements of theorems, excluding both context and proofs. %
Contextual information is extracted separately from $X$ using rule-based methods. %

The prompt given to the LLM specifies that the generated output must begin with \texttt{theorem} %
and end with \texttt{:= by}, %
and it is required to output a list of strings (\texttt{list[str]}) to ensure structured formatting. %
This setup restricts the model from producing contextual definitions or proofs. %

The system prompt used for theorem generation is as follows: %

\begin{figure}[h]
\begin{quote}
\textit{``Please generate new theorems in Lean 4 format that are similar but not identical to each theorem provided in the text as many as possible. For each theorem in the text, generate a corresponding new theorem with slight variations in content. Do not include proofs, annotations, or imports. The new theorems begin with `\texttt{```lean theorem}', not any annotations. They should end with `\texttt{:= by```}'. Additionally, please use standard mathematical symbols (e.g., $\forall$, $\exists$, $\sqrt{}$) instead of Unicode escape sequences (e.g., \textbackslash u2200).''}
\end{quote}
\caption{System prompt used for theorem generation with LLMs.}
\label{fig:system_prompt}
\end{figure}

A key design choice in our prompt is the inclusion of the phrase ``as many as possible''. %
This instruction is crucial for ensuring consistent generation volume regardless of the input file size. %
Without this directive, we observed that the number of generated conjectures tends to be influenced %
by the number of theorems in the source Mathlib file, %
potentially resulting in only a single conjecture when the input file contains just one theorem. %
The ``as many as possible'' instruction enables the generation of multiple diverse conjectures %
even from files with limited theorem content, thereby maximizing the utility of LeanConjecturer. %

To prevent syntactic issues, we apply post-processing to strip any prefixes such as modifiers %
(e.g., \texttt{protected}) %
and annotations (e.g., \texttt{@[simp]}) from the start of each theorem, %
and to remove the proof part following \texttt{:= by}. %

Since the LLM-generated outputs do not include necessary imports or contextual definitions, %
we programmatically augment each generated conjecture with appropriate imports and context. %
All generated theorems are prepended with the imports %
\begin{verbatim}
import Mathlib
import Aesop
\end{verbatim}
, allowing access to the full Mathlib library and automated proof tactics \texttt{aesop}. %
For the context, we extract global variable declarations and open any namespaces used in $X$, %
applying them to the generated conjectures. %

This hybrid approach—rule-based extraction of import/context and LLM-based generation of theorem statements—%
reduces syntactic errors and improves compatibility with the Lean 4 proof assistant. %

\subsection{Evaluation}
After generating conjectures, we perform a evaluation process in our pipeline. %
The evaluation of LLM-generated conjectures serves two crucial purposes. %
First, it helps filter out conjectures that should not be sent to theorem provers, %
thereby reducing unnecessary computational overhead. Second, %
by incorporating high-quality conjectures into the context for subsequent generation, %
we can improve the overall quality of generated conjectures through an iterative generation process.

We evaluate each generated theorem from the following perspectives. %
\begin{itemize}
\item \textbf{Syntactic Validity:} We check whether the theorem parses correctly in Lean 4. %
      If replacing the proof with \texttt{sorry} results in a single warning \texttt{declaration uses 'sorry'}, %
      we consider the syntax valid. %
\item \textbf{Novelty:} We assess whether the theorem already exists in Mathlib. %
      This is done by using Lean's \texttt{exact?} command to search for existing theorems %
      that could trivially prove the statement. %
      Importantly, we also check novelty against previously generated conjectures in our context, %
      which helps filter out duplicate or similar conjectures even when the LLM generates them repeatedly. %
      This comprehensive novelty check ensures that each generated conjecture brings new mathematical insight. %
\item \textbf{Non-Triviality:} We test whether the theorem can be automatically proven %
      using the \texttt{aesop} tactic. If \texttt{aesop} fails, %
      the conjecture is considered potentially non-trivial. %
      While non-trivial conjectures are valuable for evaluating theorem generation quality, %
      we retain conjectures that can be proven by \texttt{aesop} for use in GRPO-based reinforcement learning. %
      This approach allows us to maximize the training data available for improving theorem proving capabilities. %
\end{itemize}

It is important to note that when a conjecture cannot be proven by either \texttt{exact?} or \texttt{aesop}, %
its truth value remains undetermined. Such conjectures may include statements that are either false %
or require more sophisticated proof techniques beyond the capabilities of automated tools. %

\subsection{Iteration}
Our pipeline employs an iterative approach to maximize the generation of novel conjectures from a single Mathlib file. %
After evaluating the initial set of generated conjectures, %
we collect the syntactically valid and novel ones into a new file. %
This aggregated file then serves as the inspiration for the next round of generation. %

The iteration continues until either no new novel conjectures are generated or %
a predefined maximum number of iterations is reached. %
This iterative process allows us to explore the mathematical space more thoroughly %
and generate a diverse set of conjectures that maintain the style and complexity of the original Mathlib content. %

By adding generated theorems to the context in each iteration, %
we can assess novelty not only against Mathlib but also against previously generated conjectures, %
enabling us to filter out duplicates. %
This helps prevent a common issue with LLMs where they tend to repeat similar patterns, %
ensuring each generated conjecture brings new mathematical insight. %

\begin{figure}[h]
    \centering
    \includegraphics[width=0.5\textwidth]{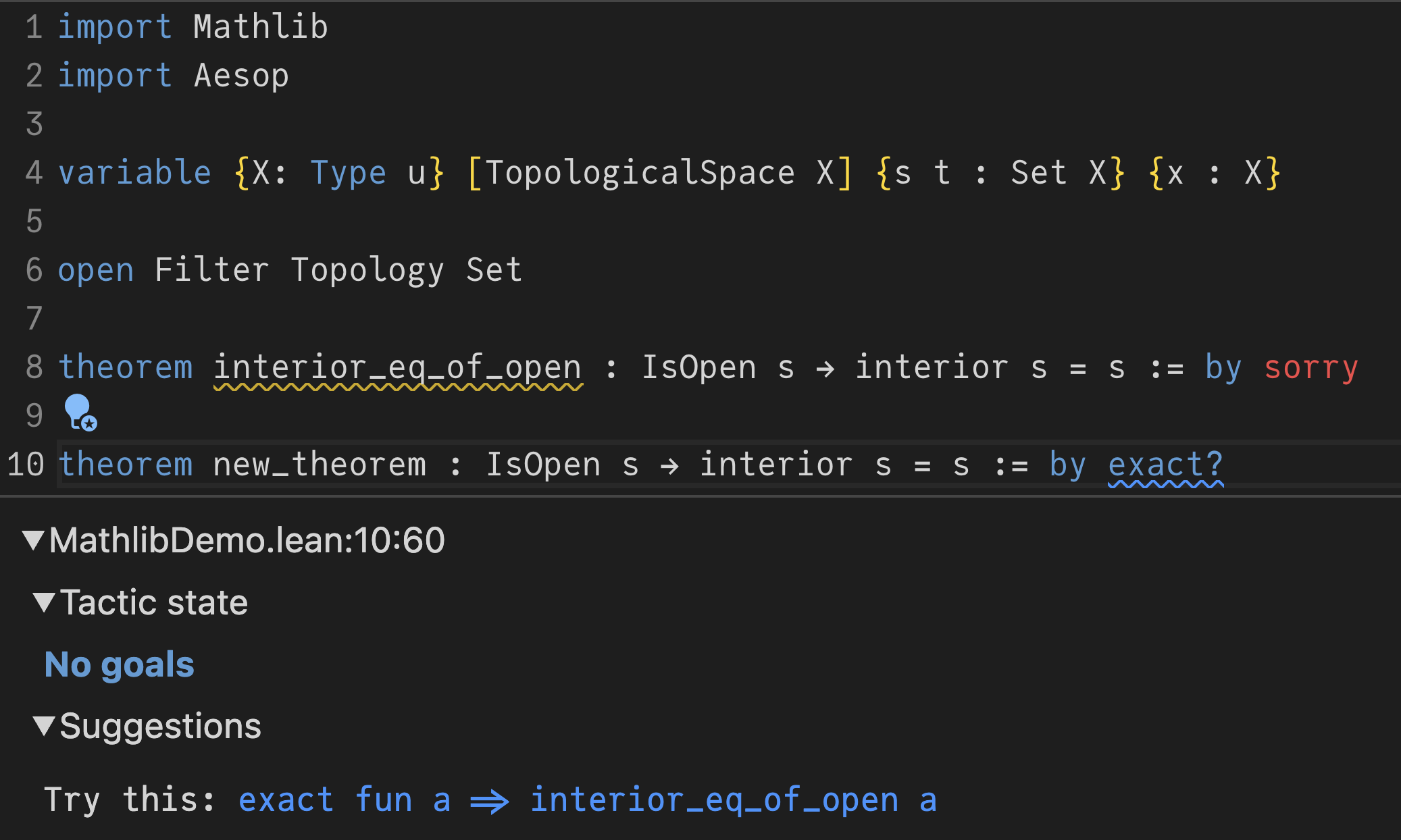}
    \caption{Illustration of novelty assessment using \texttt{exact?} command with context. By including previously generated conjectures in the context, we can effectively detect and filter out duplicate or similar conjectures, ensuring the generation of truly novel mathematical statements.}
    \label{fig:exact_incontext}
\end{figure}

The iteration process terminates when no new syntactically valid and novel conjectures are generated. %
However, the ``as many as possible'' instruction in our system prompt (Figure \ref{fig:system_prompt}) %
plays a crucial role in maintaining the robustness of the iteration process. %
This directive ensures that even when a file contains only a single theorem, %
the system generates multiple conjectures, making the iteration process less likely to terminate prematurely. %

\subsection{Verification}
Through this iterative process, we accumulate conjectures that are syntactically valid, novel, %
and cannot be proven by the \texttt{aesop} tactic. %
It is important to note that conjectures that are syntactically valid, novel, and cannot be proven by \texttt{aesop} %
may include false propositions. %
To definitively establish their truth, formal proofs must be provided. %
We accomplish this verification process using DeepSeek Prover-V2, %
an open-source large language model that achieves an impressive 88.9\% success rate on the MiniF2F benchmark. %
To enable DeepSeek Prover-V2 to effectively handle Mathlib-style problems, %
we also conduct reinforcement learning using the conjectures prepared by LeanConjecturer. %

\subsection{Reinforcement Learning}
To enhance the theorem proving capabilities, we apply Group Relative Policy Optimization (GRPO) %
for reinforcement learning. %
GRPO requires both a reward function and a well-defined problem set for reinforcement learning optimization of LLM. %
We use DeepSeek Prover-V2 as our theorem prover. %
DeepSeek Prover-V2 attempts to provide proofs for the novel theorems generated by LeanConjecturer. %
We employ a binary reward function that returns 1 for successful proofs and 0 for failed attempts. %

LeanConjecturer is specifically designed to produce theorems similar to existing ones, %
making it ideal for enhancing proof capabilities in targeted mathematical domains. %

For our reinforcement learning experiments, we utilize the conjectures prepared through LeanConjecturer. %
In this study, we focus exclusively on conjectures related to properties of interior and closure operations %
in topological spaces, generated from a single seed file (See \url{https://github.com/auto-res/LeanConjecturer/blob/main/InterClosureExercise.lean}). %
Through this reinforcement learning process, we measure how DeepSeek Prover-V2 %
improves its ability to solve problems involving closure and interior operations. %
Additionally, we evaluate the model's enhanced proof capabilities on concepts related to %
semi-open sets in topological spaces, providing insights into the transferability %
of the learned skills to related mathematical concepts. %

We also investigated whether repeated GRPO training epochs on problems involving %
interior and closure operations in topological spaces lead to further improvements %
in proof-finding capabilities, allowing us to assess the scalability and effectiveness %
of extended reinforcement learning training. %

\section{Results}

\subsection{Experimental Setup}

Our experiments were conducted using Lean 4 (v4.18.0-rc1) for evaluating generated conjectures and validating proofs. %

To evaluate the performance of our conjecture generation system, we selected 40 files from Mathlib %
as seed files and executed the generation/evaluation/iteration process as described in the Methods section. %

In order to evaluate whether GRPO-based reinforcement learning %
could effectively improve theorem proving capabilities for specific problems, %
we used the file \texttt{InteriorClosureExercise.lean} as our seed file. %
This file contains properties about interior and closure operations %
in topological spaces. %

For the LLM configuration, we utilized OpenAI's o3 model, with a maximum iteration count set to 15. %
The GRPO training setup employed the DeepSeek Prover V2 7B model %
with a binary reward function for successful proofs. %
The reward function implementation and additional hyperparameter configurations %
can be found in our forked repository\footnote{\url{https://github.com/auto-res/open-r1.git}}.

The experiments were performed on a high-performance computing environment equipped with 8 NVIDIA A100 GPUs, %
an AMD EPYC 7742 64-Core Processor, and 2.0 TB of memory, providing the necessary computational resources %
for large-scale proof generation and reinforcement learning.

\subsection{Conjecture Generation Experiments}

We conducted comprehensive experiments using 40 selected Mathlib files as seed files, %
with iterative conjecture generation using a maximum iteration count of 15. %
Using our pipeline, a total of 12289 conjectures were generated across the 40 selected files. %
Of these, 10950 conjectures were free from syntax errors and could be parsed by Lean 4. %
Among them, 4130 were identified as conjectures that were not already present in Mathlib %
and 3776 could not be automatically proven by \texttt{aesop}. %
On average, 103.25 novel conjectures are generated from a single file through our iterative process. %

The following table summarizes the distribution of generated conjectures across various categories: %

\begin{table}[h]
  \centering
  \caption{Evaluation of generated theorems}
  \label{tab:eval}
  \begin{tabular}{lcccc}
      \toprule
      \textbf{Total} & \textbf{Valid} & \textbf{Novel} & \textbf{Non-Trivial}\\
      \midrule
      12289 & 10950  & 4130 & 3776\\
      \bottomrule
    \end{tabular}
\end{table}

It is worth noting that our experiments used a maximum iteration count of 15, %
and among the 40 files tested, 25 files reached this maximum iteration limit. %
This observation suggests that increasing the maximum iteration count %
could generate more conjectures, %
as these files were still actively producing novel conjectures when the iteration process was terminated.

The inclusion of the ``as many as possible'' instruction %
in the prompt proved crucial for maintaining consistent generation volume regardless of the input file size. %
Without this directive, we observed that the number of generated conjectures tended to be influenced %
by the number of theorems in the source Mathlib file. %
For instance, when processing the \texttt{Mathlib.Algebra.Group.Commutator} library from version 4.18.0-rc1, %
which contains only one theorem, the iteration process would terminate prematurely without generating multiple conjectures. %
However, with the ``as many as possible'' instruction, the iteration process continued, %
resulting in the generation of 412 theorems from this file alone, with 108 of them being syntactically valid %
and non-trivial (unprovable by \texttt{aesop}). %

\begin{table}[h]
  \centering
  \caption{Comparison of generation results with seed file \texttt{Mathlib.Algebra.Group.Commutator}}
  \label{tab:comparison}
  \begin{tabular}{lccc}
      \toprule
      \textbf{Prompt} & \textbf{Total} & \textbf{Valid} & \textbf{Non-Trivial} \\
      \midrule
      Without ``as many & 1 & 0 & 0 \\
      as possible'' & & & \\
      With ``as many & 412 & 392 & 108 \\
      as possible'' & & & \\
      \bottomrule
  \end{tabular}
\end{table}

\subsection{GRPO Learning Experiments}

To further enhance theorem proving capabilities, we conducted experiments using GRPO %
(Group Relative Policy Optimization) on a dataset of 192 conjectures in general topology, %
focusing on properties related to interior and closure operations. %
These conjectures were generated by LeanConjecturer without manual filtering, %
thus including both true and false statements. %
Our goal was to improve the model's performance in specific mathematical domains %
through targeted additional training. %

We first compared the performance of DeepSeekProver-V1.5-RL and DeepSeekProver-V2-7B, %
and proceeded with additional training on the latter model which showed better results. %
For each problem, we performed 128 sampling attempts, %
considering a problem solved if at least one attempt was successful. %

The per-proof success rate showed consistent improvement across training epochs, %
and the per-problem success rate also increased, %
indicating that previously unsolvable problems became solvable after training. %
The results are shown in Table \ref{tab:grpo}. %

\begin{table}[h]
    \centering
    \caption{Proof success rates in the training dataset}
    \label{tab:grpo}
    \begin{tabular}{lcc}
        \toprule
        \textbf{Model} & \textbf{Proof Rate} & \textbf{Problem Rate} \\
        \midrule
        DSPV1.5-RL & 982/24576 & 43/192 \\
        DSPV2-7B & 2285/24576 & 47/192 \\
        +GRPO (1 epoch) & 2306/24576 & 48/192 \\
        +GRPO (10 epochs) & 3657/24576 & 49/192 \\
        \bottomrule
    \end{tabular}
\end{table}

The training for 10 epochs took approximately 16 hours. %

After 10 epochs, we added 121 semi-open set conjectures, that were also generated by LeanConjecturer, to the training dataset %
and continued training for an additional 24 epochs to further improve performance. %
\begin{table}[h]
    \centering
    \caption{Proof success rates in the training dataset with semi-open set conjectures after additional 24 epochs}
    \label{tab:grpo2}
    \begin{tabular}{lcc}
        \toprule
        \textbf{Model} & \textbf{Proof Rate} & \textbf{Problem Rate} \\
        \midrule
        10+24 epochs & 5307/24576 & 50/192 \\
        \bottomrule
    \end{tabular}
\end{table}

\begin{figure}[h]
    \centering
    \includegraphics[width=0.5\textwidth]{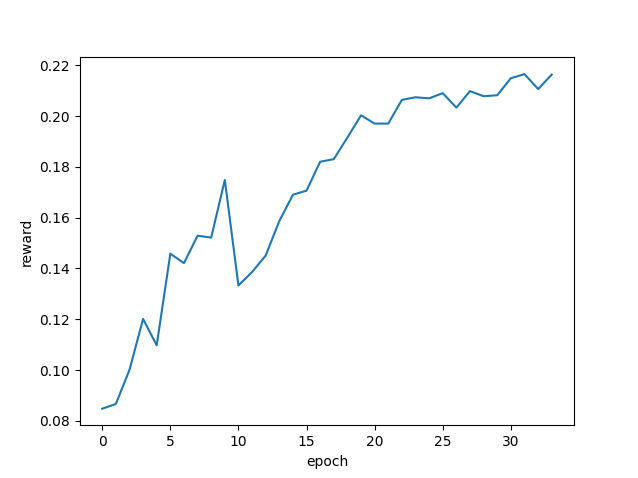}
    \caption{Learning curve of GRPO training over 34 epochs. The graph shows the improvement in proof success rates and problem success rates as training progresses.}
    \label{fig:learning_curve}
\end{figure}

The progression of GRPO rewards throughout the training process is shown in Figure~\ref{fig:learning_curve}. %

To investigate the generalization capabilities of GRPO training, %
we evaluated the model's performance on a separate dataset of 14 conjectures %
involving alpha-open sets and related concepts. %
The results showed that the model's performance on these new concepts remained comparable to the base model: %

\begin{table}[h]
  \centering
  \caption{Performance on alpha-open set conjectures}
  \label{tab:semiopen}
  \begin{tabular}{lcc}
      \toprule
      \textbf{Model} & \textbf{Proof Rate} & \textbf{Problem Rate} \\
      \midrule
      DSPV2-7B & 466/1792 & 9/14 \\
      +GRPO (1 epoch) & 443/1792 & 9/14 \\
      +GRPO (10 epochs) & 454/1792 & 8/14 \\
      10+24 epochs & 536/1792 & 9/14 \\
      \bottomrule
  \end{tabular}
\end{table}

The problem rate did not exhibit a clear upward trend, %
but the proof rate shows an improvement. %

\subsection{Verified Conjectures}

Through our pipeline, we generated and verified a collection of theorems related to topological spaces, %
particularly focusing on semi-open, alpha-open, and pre-open sets. %
These concepts extend the traditional notion of open sets in topology, %
providing a richer framework for studying topological properties. %
A set is called \textit{semi-open} if it is a subset of the closure of its interior, %
\textit{alpha-open} if it is a subset of the interior of the closure of its interior, %
and \textit{pre-open} if it is a subset of the interior of its closure. %

The generated library includes several interesting and non-trivial theorems. %
For instance, we proved that the closure of a pre-open set is semi-open, which, %
while straightforward to prove, offers an interesting connection between these different types of sets. %
Other notable results include:

\begin{itemize}
    \item The union of two semi-open sets is semi-open
    \item The union of two alpha-open sets is alpha-open
    \item The union of two pre-open sets is pre-open
    \item The closure of a pre-open set is semi-open
\end{itemize}

The generation process revealed some interesting patterns in the LLM's behavior. %
When generating conjectures with previously proven theorems in the context, %
the model tended to produce similar patterns of theorems. %
However, when including all conjectured statements in the context, %
the model showed more diverse and sometimes unexpected variations. %
This suggests that the context management strategy influences the diversity and creativity of the generated conjectures. %

We also observed that the model sometimes generated redundant or unnecessary assumptions in its proofs. %
For example, some proofs continued with additional steps even after the theorem was already proven, %
using tactics such as `<;>` to avoid errors. This behavior likely stems %
from the model's inability to determine when a proof is complete during generation. %
This suggests that an interactive, step-by-step proof generation approach might produce more readable and efficient proofs.

The generated library demonstrates that LeanConjecturer can produce meaningful mathematical content %
beyond simple variations of existing theorems. The theorems maintain mathematical rigor %
while exploring interesting properties of topological spaces. %
This success in generating verified conjectures in a specific mathematical domain (topology) suggests that %
our approach could be extended to other areas of mathematics, potentially leading to new mathematical discoveries. %

\section{Discussion}

\subsection{LeanConjecturer Effectiveness and Contributions}

Our work presents a pipeline for automatically generating mathematical conjectures in Lean 4, %
demonstrating its utility in addressing the data scarcity issue for training large language models in formal theorem proving. %
The hybrid approach, combining rule-based context extraction with LLM-based theorem statement generation, %
mitigates common problems such as hallucinated imports and syntactic errors that plague direct LLM generation. %
The iterative nature of LeanConjecturer allows for continuous creation of novel conjectures, %
with the generation of non-trivial conjectures (3776 out of 10950 valid ones) %
indicating that our system can pose challenges for state-of-the-art provers. %
Furthermore, given that Mathlib contains approximately 6000 files compared to the 40 files used in our experiments, %
and considering that 25 out of 40 files reached the maximum iteration count of 15, %
increasing the maximum iteration count and extending our approach to the entire Mathlib library %
could yield substantially more non-trivial conjectures for training theorem proving systems.

\subsection{Reinforcement Learning Insights}

The GRPO experiments reveal insights about reinforcement learning in theorem proving. %
The progressive improvement in proof success rates (from 2285 to 5307 successful proofs out of 24576 attempts) %
demonstrates that targeted training on domain-specific conjectures can enhance theorem proving capabilities. %
However, the modest improvement in problem success rates (from 47 to 50 out of 192 problems) %
suggests that while the model becomes more proficient at proving specific types of theorems, %
it may not automatically generalize to entirely new problem types within the same domain. %
The generalization results on alpha-open set conjectures indicate that while GRPO improves performance on the training distribution, %
its ability to generalize to novel mathematical concepts requires more sophisticated training strategies. %

\subsection{Limitations and Future Work}

Several limitations of our current pipeline warrant discussion. %
The primary challenge remains the automated verification of the truthfulness of generated conjectures, %
which still relies on theorem provers such as DeepSeek Prover-V2. %
Furthermore, the quality of generated theorems is inherently tied to the quality and diversity %
of the initial Mathlib seed files, potentially introducing bias if the seeds are not sufficiently representative. %
The computational cost of 16 hours for 10 epochs of GRPO training also poses scalability challenges. %

Future work should focus on refining the conjecture generation process to produce deeper %
and more insightful mathematical statements, possibly by incorporating techniques %
for guiding the LLM towards unexplored areas of mathematical theory. %
Exploring advanced reinforcement learning techniques with more sophisticated reward functions %
could lead to more robust and generalized improvements in theorem proving capabilities. %
The potential for combining our conjecture generation approach with self-play systems %
such as STP represents a direction for future research, %
potentially creating hybrid systems that leverage the strengths of both approaches.

\section*{Acknowledgements}
This work was supported by JSPS KAKENHI 24K21316, %
JST BOOST JPMJBY24E2, JST CREST JPMJCR2015, KAKENHI PROJECT 24K16077, %
Advanced General Intelligence for Science Program(AGIS), %
the RIKEN TRIP initiative %
and JST Moonshot %
R\&D Program JPMJMS2236.

\bibliography{references}
\bibliographystyle{icml2025}

\newpage
\appendix
\onecolumn

\section{Generated and Verified Conjectures}
\label{app:generated_and_verified_conjectures}

The followings are examples of generated and verified conjectures.
\begin{lstlisting}
import Mathlib
import Aesop

namespace Topology

variable {X : Type*} [TopologicalSpace X]

/-- A set is semi-open if it is a subset of the closure of its interior. -/
def SemiOpen (A : Set X) : Prop :=
  A ⊆ closure (interior A)

/-- A set is alpha-open if it is a subset of the interior of the closure of its interior. -/
def AlphaOpen (A : Set X) : Prop :=
  A ⊆ interior (closure (interior A))

/-- A set is preopen if it is a subset of the interior of its closure. -/
def PreOpen (A : Set X) : Prop :=
  A ⊆ interior (closure A)
\end{lstlisting}
\begin{lstlisting}

theorem semi_open_union {A B : Set X} (hA : SemiOpen A) (hB : SemiOpen B) 
  : SemiOpen (A ∪ B) :=  by
  intro x hx
  have h1 : x ∈ A ∪ B := hx
  cases h1 with
  | inl h1 =>
    -- If x is in A, then x is in the closure of the interior of A
    have h2 : x ∈ closure (interior A) := hA h1
    have h3 : closure (interior A) ⊆ closure (interior (A ∪ B)) := by
      apply closure_mono
      apply interior_mono
      apply Set.subset_union_left
    exact h3 h2
  | inr h1 =>
    -- If x is in B, then x is in the closure of the interior of B
    have h2 : x ∈ closure (interior B) := hB h1
    have h3 : closure (interior B) ⊆ closure (interior (A ∪ B)) := by
      apply closure_mono
      apply interior_mono
      apply Set.subset_union_right
    exact h3 h2

\end{lstlisting}
\begin{lstlisting}
theorem open_set_is_alpha_open (A : Set X) (h : IsOpen A) : AlphaOpen A :=  by
  intro x hx
  -- Since A is open, the interior of A is A itself.
  have h1 : interior A = A := h.interior_eq
  -- Simplify the expression using the fact that the interior of A is A.
  simp_all [AlphaOpen]
  -- Since x is in A and A is open, x is in the interior of the closure of A.
  <;> apply interior_mono (subset_closure)
  <;> simp_all [AlphaOpen]

theorem closure_pre_open_is_semi_open (A : Set X) (h : PreOpen A) 
  : SemiOpen (closure A) :=  by
  -- We need to show that the closure of A is semi-open.
  -- By definition, a set is semi-open if it is a subset of the closure of its interior.
  -- Given that A is preopen, we know A ⊆ interior (closure A).
  -- Taking the closure of both sides, we get closure A ⊆ closure (interior (closure A)).
  -- Since closure is idempotent, closure (interior (closure A)) = closure A.
  -- Therefore, closure A ⊆ closure A, which is trivially true.
  simp_all [SemiOpen, PreOpen, closure_mono, interior_mono, subset_closure]
  <;> tauto

\end{lstlisting}
\begin{lstlisting}
theorem closure_subset_of_semi_open {A : Set X} (h : SemiOpen A) 
  : closure A ⊆ closure (interior A) := 
  calc
    closure A ⊆ closure (closure (interior A)) := by
      -- Since A is semi-open, A ⊆ closure (interior A).
      -- Therefore, closure A ⊆ closure (closure (interior A)).
      exact closure_mono h
    _ = closure (interior A) := by
      -- By the properties of closures, closure (closure (interior A)) = closure (interior A).
      rw [closure_closure]

theorem preopen_empty_set : PreOpen (∅ : Set X) :=  by
  intro x hx
  -- We need to show that an empty set satisfies the condition of being preopen.
  -- The condition is that for every x in ∅, there exists a neighborhood of x contained in ∅.
  -- Since there are no elements in ∅, this condition is vacuously true.
  simp at hx

\end{lstlisting}
\begin{lstlisting}
theorem semi_open_empty : SemiOpen (∅ : Set X) :=  by
  -- Simplify the goal using the definition of SemiOpen and the properties of ∅.
  simp [SemiOpen, closure_empty]
  -- The empty set is a subset of itself, and the closure of the empty set is the empty set.
  -- Therefore, the condition ∅ ⊆ closure (interior ∅) holds because closure ∅ = ∅.
  <;> aesop

\end{lstlisting}
\begin{lstlisting}
theorem interior_union_preopen {A B : Set X} (hA : PreOpen A) (hB : PreOpen B) 
  : interior (A ∪ B) ⊆ interior (closure (A ∪ B)) :=  by
  -- Use the fact that the interior of a union is contained in the union of the interiors
  apply interior_mono
  -- The union of A and B is contained in the closure of the union of A and B
  apply subset_closure
  -- The union of A and B is the union of A and B
  <;> simp_all [PreOpen, closure_union]
  -- Use the fact that the closure of the union is the union of the closures
  <;> apply interior_mono
  <;> simp_all [PreOpen, closure_union]
  <;> apply subset_closure
  <;> simp_all [PreOpen, closure_union]
  <;> apply interior_mono
  <;> simp_all [PreOpen, closure_union]
  <;> apply subset_closure
  <;> simp_all [PreOpen, closure_union]
  <;> apply interior_mono
  <;> simp_all [PreOpen, closure_union]
  <;> apply subset_closure
  <;> simp_all [PreOpen, closure_union]
  <;> apply interior_mono
  <;> simp_all [PreOpen, closure_union]
  <;> apply subset_closure
  <;> simp_all [PreOpen, closure_union]
  <;> apply interior_mono
  <;> simp_all [PreOpen, closure_union]
  <;> apply subset_closure
  <;> simp_all [PreOpen, closure_union]
  <;> apply interior_mono
  <;> simp_all [PreOpen, closure_union]
  <;> apply subset_closure
  <;> simp_all [PreOpen, closure_union]
  <;> apply interior_mono
  <;> simp_all [PreOpen, closure_union]
  <;> apply subset_closure
  <;> simp_all [PreOpen, closure_union]

\end{lstlisting}
\begin{lstlisting}
theorem interior_alpha_open_subset_alpha_open (A : Set X) 
  : AlphaOpen A → interior A ⊆ interior (closure (interior A)) :=  by
  intro h
  exact fun x hx => h (interior_subset hx)

theorem semi_open_interior_alpha_open (A : Set X) : SemiOpen A → AlphaOpen (interior A) :=  by
  intro h
  simp_all [SemiOpen, AlphaOpen, interior_mono, interior_interior]
  <;>
  exact?

theorem pre_open_set_is_pre_open (A : Set X) (h : PreOpen A) : PreOpen (interior A) :=  by
  simp_all [PreOpen, interior_mono, interior_interior]
  <;> apply interior_maximal <;> simp_all [interior_subset, subset_closure]
  <;> apply interior_maximal <;> simp_all [interior_subset, subset_closure]

\end{lstlisting}
\begin{lstlisting}
theorem alpha_open_interior_closure (A : Set X) (h : AlphaOpen A) 
  : AlphaOpen (interior (closure A)) :=  by
  simp_all [AlphaOpen, interior_mono, closure_mono, interior_interior]
  -- Use the fact that the interior of a set is a subset of the interior of its closure.
  -- Use the fact that the closure of a set is a subset of the closure of its interior.
  -- Combine these facts to show that the interior of the closure of a set 
  -- is a subset of the interior of the closure of its interior.
  <;> apply interior_maximal <;> simp_all [interior_subset, subset_closure]
  -- Use the fact that the interior of a set is the largest open set contained in the set.
  -- Use the fact that the closure of a set is the smallest closed set containing the set.
  -- Combine these facts to show that the interior of the closure of a set 
  -- is a subset of the interior of the closure of its interior.
  <;> apply closure_mono <;> simp_all [interior_subset, subset_closure]
  -- Use the fact that the closure of a set is the smallest closed set containing the set.
  -- Use the fact that the interior of a set is the largest open set contained in the set.
  -- Combine these facts to show that the interior of the closure of a set 
  -- is a subset of the interior of the closure of its interior.
  <;> apply interior_mono <;> simp_all [interior_subset, subset_closure]
  -- Use the fact that the interior of a set is the largest open set contained in the set.
  -- Use the fact that the closure of a set is the smallest closed set containing the set.
  -- Combine these facts to show that the interior of the closure of a set 
  -- is a subset of the interior of the closure of its interior.
  <;> apply closure_mono <;> simp_all [interior_subset, subset_closure]

\end{lstlisting}
\begin{lstlisting}
theorem alpha_open_empty_set : AlphaOpen (∅ : Set X) :=  by
  intro x hx
  simp_all [AlphaOpen, interior_empty, closure_empty]
  <;> aesop

theorem interior_of_alpha_open_is_alpha_open (A : Set X) (h : AlphaOpen A) 
  : AlphaOpen (interior A) :=  by
  simp_all [AlphaOpen, interior_mono, interior_interior]
  <;>
    apply interior_maximal <;>
    simp_all [interior_subset, subset_closure, closure_mono, interior_mono] <;>
    apply closure_mono <;>
    simp_all [interior_subset, subset_closure, closure_mono, interior_mono] <;>
    apply interior_subset
  <;>
    simp_all [interior_subset, subset_closure, closure_mono, interior_mono]
  <;>
    apply closure_mono <;>
    simp_all [interior_subset, subset_closure, closure_mono, interior_mono]
  <;>
    apply interior_subset

\end{lstlisting}
\begin{lstlisting}
theorem finitary_preservation_of_semi_open (U : Set (Set X)) (h : ∀ A ∈ U, SemiOpen A)  
  : SemiOpen (⋃₀ U) :=  by
  intro x hx
  simp only [Set.mem_sUnion, exists_prop] at hx
  rcases hx with ⟨V, hVU, hxV⟩
  have hV : SemiOpen V := h V hVU
  have h1 : V ⊆ closure (interior V) := hV
  have h2 : x ∈ closure (interior V) := h1 hxV
  exact closure_mono (interior_mono (Set.subset_sUnion_of_mem hVU)) h2

theorem semi_open_of_open (A : Set X) (h : IsOpen A) : SemiOpen A :=  by
  intro x hx
  have h₁ : x ∈ closure (interior A) := by
    apply subset_closure
    exact h.interior_eq.symm ▸ hx
  exact h₁

theorem semi_open_interior_subset_interior (A : Set X) 
  : SemiOpen A → interior A ⊆ interior (closure A) :=  by
  intro h
  have h₁ : interior A ⊆ interior (closure A) := by
    -- Use the fact that the interior of a set is a subset of the interior of its closure
    exact interior_mono (subset_closure)
  exact h₁

\end{lstlisting}
\begin{lstlisting}
theorem alpha_open_union {A B : Set X} (hA : AlphaOpen A) (hB : AlphaOpen B) 
  : AlphaOpen (A ∪ B) :=  by
  intro x hx
  have h₁ : x ∈ A ∪ B := hx
  cases h₁ with
  | inl h₁ =>
    -- If x is in A, then by AlphaOpen property of A,
    -- x is in the interior of the closure of the interior of A.
    have h₂ : x ∈ interior (closure (interior A)) := hA h₁
    have h₃ : interior (closure (interior A)) ⊆ interior (closure (interior (A ∪ B))) := by
      apply interior_mono
      apply closure_mono
      apply interior_mono
      apply Set.subset_union_left
    exact h₃ h₂
  | inr h₁ =>
    -- If x is in B, then by AlphaOpen property of B, 
    -- x is in the interior of the closure of the interior of B.
    have h₂ : x ∈ interior (closure (interior B)) := hB h₁
    have h₃ : interior (closure (interior B)) ⊆ interior (closure (interior (A ∪ B))) := by
      apply interior_mono
      apply closure_mono
      apply interior_mono
      apply Set.subset_union_right
    exact h₃ h₂

\end{lstlisting}
\begin{lstlisting}
theorem closure_interior_subset_closure (A : Set X) : closure (interior A) ⊆ closure A :=  by
  apply closure_mono
  exact interior_subset

theorem pre_open_closure_eq_closure (A : Set X) (h : PreOpen A) 
  : closure A = closure (interior (closure A)) :=  by
  have h₁ : A ⊆ interior (closure A) := h
  have h₂ : closure A ⊆ closure (interior (closure A)) := closure_mono h₁
  have h₃ : closure (interior (closure A)) ⊆ closure A := by
    apply closure_minimal
    · apply interior_subset
    · exact isClosed_closure
  exact le_antisymm h₂ h₃

theorem closure_preopen_subset (A : Set X) : closure (interior A) ⊆ closure (closure A) :=  by
  have h₁ : interior A ⊆ closure A := interior_subset_closure
  have h₂ : closure (interior A) ⊆ closure (closure A) := closure_mono h₁
  exact h₂

theorem alpha_open_subset_open_closure (A : Set X) (h : AlphaOpen A) 
: A ⊆ interior (closure A) :=  by
  have h₁ : A ⊆ interior (closure (interior A)) := h
  have h₂ : interior (closure (interior A)) ⊆ interior (closure A) 
    := interior_mono (closure_mono (interior_subset))
  exact h₁.trans h₂

\end{lstlisting}
\begin{lstlisting}
theorem interior_of_closure_is_pre_open {A : Set X} : PreOpen (interior (closure A)) :=  by
  -- We need to show that the interior of the closure of A is preopen.
  -- This means we need to show that the interior of the closure of A 
  -- is a subset of the interior of its closure.
  simp [PreOpen]
  -- Simplify the definition of preopen sets to show that the interior of the closure of A 
  -- is indeed a subset of the interior of its closure.
  <;> apply interior_maximal
  -- Apply the property that the interior of a set is the largest open subset of that set.
  <;> simp [interior_subset, subset_closure]
  -- Simplify the set inclusions and use the fact that the interior of a set 
  -- is a subset of that set and the closure of a set contains that set.
  <;> apply interior_mono
  -- Apply the property that the interior of a set is monotonic with respect to set inclusion.
  <;> simp [subset_closure]

\end{lstlisting}
\begin{lstlisting}
theorem pre_open_union {A B : Set X} (hA : PreOpen A) (hB : PreOpen B) : PreOpen (A ∪ B) :=  by
  rw [PreOpen]
  intro x hx
  rw [Set.mem_union] at hx
  cases' hx with hx hx
  · -- Case: x ∈ A
    have h1 : x ∈ interior (closure A) := hA hx
    have h2 : interior (closure A) ⊆ interior (closure (A ∪ B)) := by
      apply interior_mono
      apply closure_mono
      apply Set.subset_union_left
    exact h2 h1
  · -- Case: x ∈ B
    have h1 : x ∈ interior (closure B) := hB hx
    have h2 : interior (closure B) ⊆ interior (closure (A ∪ B)) := by
      apply interior_mono
      apply closure_mono
      apply Set.subset_union_right
    exact h2 h1

theorem preopen_closure_subset_interior_closure (A : Set X) 
  : PreOpen (closure A) → closure A ⊆ interior (closure A) :=  by
  intro h
  simp_all [PreOpen, interior_mono]
  <;> exact?
  <;> simp_all [PreOpen, interior_mono]
  <;> exact?
  <;> simp_all [PreOpen, interior_mono]
  <;> exact?

\end{lstlisting}
\begin{lstlisting}
theorem open_set_is_pre_open (A : Set X) (h : IsOpen A) : PreOpen A :=  by
  intro x hx
  have h1 : A ⊆ interior (closure A) := by
    apply interior_maximal
    · exact subset_closure
    · exact h
  exact h1 hx

theorem alpha_open_implies_semi_open {A : Set X} (h : AlphaOpen A) : SemiOpen A :=  by
  intro x hx
  have h1 : x ∈ closure (interior A) := by
    apply interior_subset
    exact h hx
  exact h1
\end{lstlisting}


\end{document}